\documentclass[letterpaper]{article} 
\usepackage{aaai23}  
\usepackage{times}  
\usepackage{helvet}  
\usepackage{courier}  
\usepackage[hyphens]{url}  
\usepackage{graphicx} 
\usepackage{subfigure}
\usepackage{natbib}  
\usepackage{caption} 
\usepackage{algorithm}
\usepackage{algorithmic}
\usepackage{amsmath}
\usepackage{amssymb}
\usepackage{booktabs}
\usepackage{multirow}
\usepackage{newfloat}
\usepackage{listings}
\DeclareCaptionStyle{ruled}{labelfont=normalfont,labelsep=colon,strut=off} 
\floatstyle{ruled}
\newfloat{listing}{tb}{lst}{}
\floatname{listing}{Listing}
\title{LORE: Logical Location Regression Network for Table Structure Recognition}
\author{
    Hangdi Xing\equalcontrib\textsuperscript{\rm 1},
    Feiyu Gao\equalcontrib\textsuperscript{\rm 3},
    Rujiao Long\textsuperscript{\rm 3},
    Jiajun Bu\textsuperscript{\rm 1},
    Qi Zheng\textsuperscript{\rm 3},\\
    Liangcheng Li\textsuperscript{\rm 1},
    Cong Yao\textsuperscript{\rm 3},
    Zhi Yu\thanks{Corresponding Author.}\textsuperscript{\rm 2}
}
\affiliations{
    \textsuperscript{\rm 1}Zhejiang Provincial Key Laboratory of Service Robot, College of Computer Science, Zhejiang University \\
    \textsuperscript{\rm 2}Zhejiang Provincial Key Laboratory of Service Robot, School of Software Technology, Zhejiang University\\
    \textsuperscript{\rm 3}DAMO Academy, Alibaba Group, Hangzhou, China

    \{xinghd, bjj, liangcheng\_li, yuzhirenzhe\}@zju.edu.cn,
    feiyu.gfy@alibaba-inc.com, \\
    \{rujiao.lrj, yaocong2010\}@gmail.com,
    yongqi.zq@taobao.com
    
}

\usepackage{bibentry}

\begin{document}

\maketitle

\begin{abstract}
Table structure recognition (TSR) aims at extracting tables in images into machine-understandable formats. Recent methods solve this problem by predicting the adjacency relations of detected cell boxes, or learning to generate the corresponding markup sequences from the table images. However, they either count on additional heuristic rules to recover the table structures, or require a huge amount of training data and time-consuming sequential decoders. In this paper, we propose an alternative paradigm. We model TSR as a logical location regression problem and propose a new TSR framework called LORE, standing for LOgical location REgression network, which for the first time combines logical location regression together with spatial location regression of table cells. Our proposed LORE is conceptually simpler, easier to train and more accurate than previous TSR models of other paradigms. Experiments on standard benchmarks demonstrate that LORE consistently outperforms prior arts. Code is available at https://
github.com/AlibabaResearch/AdvancedLiterateMachinery/
tree/main/DocumentUnderstanding/LORE-TSR.
\end{abstract}

\section{Introduction}
Data in tabular format is prevalent in various sorts of documents for summarizing and presenting information. As the world is going digital, the need for parsing the tables trapped in unstructured data (e.g., images and PDF files) is growing rapidly.  Although straightforward for humans, it is challenging for automated systems due to the wide diversity of layouts and styles of tables. Table Structure Recognition (TSR) refers to transforming tables in images to machine-understandable formats, usually in logical coordinates or markup sequences. The extracted table structures are crucial for information retrieval, table-to-text generation and question answering systems, etc. 

\begin{figure}[t]
  \centering
  
    \subfigure[Adjacency relationship representations] {\includegraphics[width=0.97\linewidth]{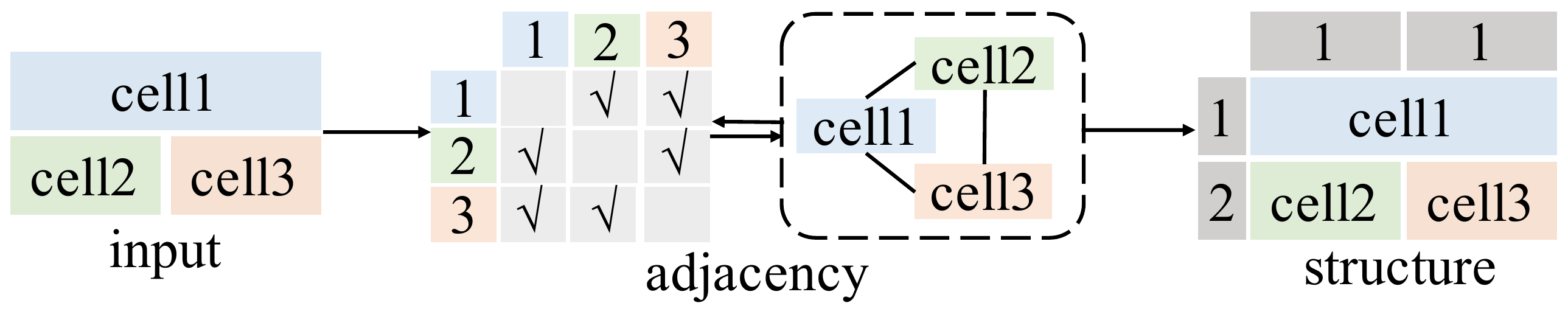}}
    
    \subfigure[Markup sequence representations] 
    {\includegraphics[width=0.97\linewidth]{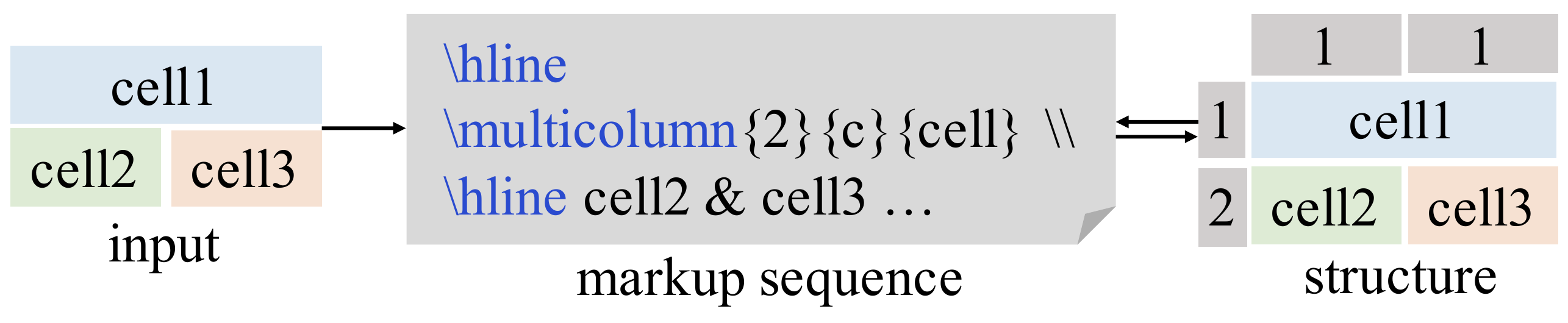}}
    
    \subfigure[Logical location representations] 
    {\includegraphics[width=0.97\linewidth]{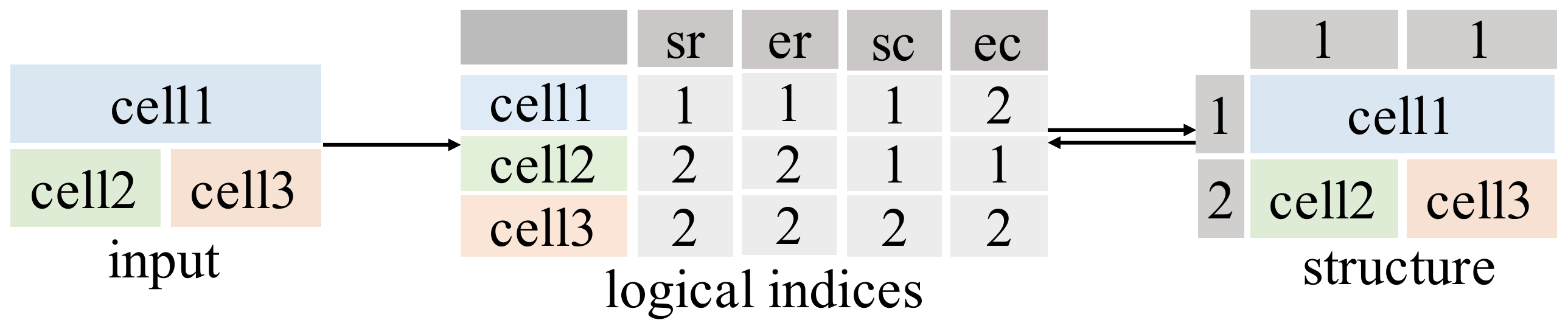}}
    
    \caption{TSR paradigms using different table-structure representations. Here, $sr$, $er$, $sc$, $ec$ refer to the starting-row, ending-row, starting-column and ending-column respectively.}
    
    \label{fig:type}
    
\end{figure}

With the development of deep learning, TSR methods have recently advanced substantially. Most deep learning-based TSR methods can be categorized into the following paradigms. The first type of models \cite{chi2019complicated, raja2020table, liu2021neural} aim at exploring the adjacency relationships between pairs of detected cells to generate intermediate results. They rely on tedious post-processings or graph optimization algorithms to reconstruct the table as logical coordinates, as depicted in Figure \ref{fig:type} (a), which would struggle with complex table structures. Another paradigm formulates TSR as a markup language sequence generation problem \cite{zhong2020image, desai2021tablex}, as shown in Figure \ref{fig:type} (b). It simplifies the TSR pipelines, but the models are supposed to redundantly learn the markup grammar from noisy sequence labels, which results in a much larger amount of training data. Besides, these models are time-consuming due to the sequential decoding process.

\begin{figure}[t]
  \centering
\includegraphics[width=0.972\columnwidth]{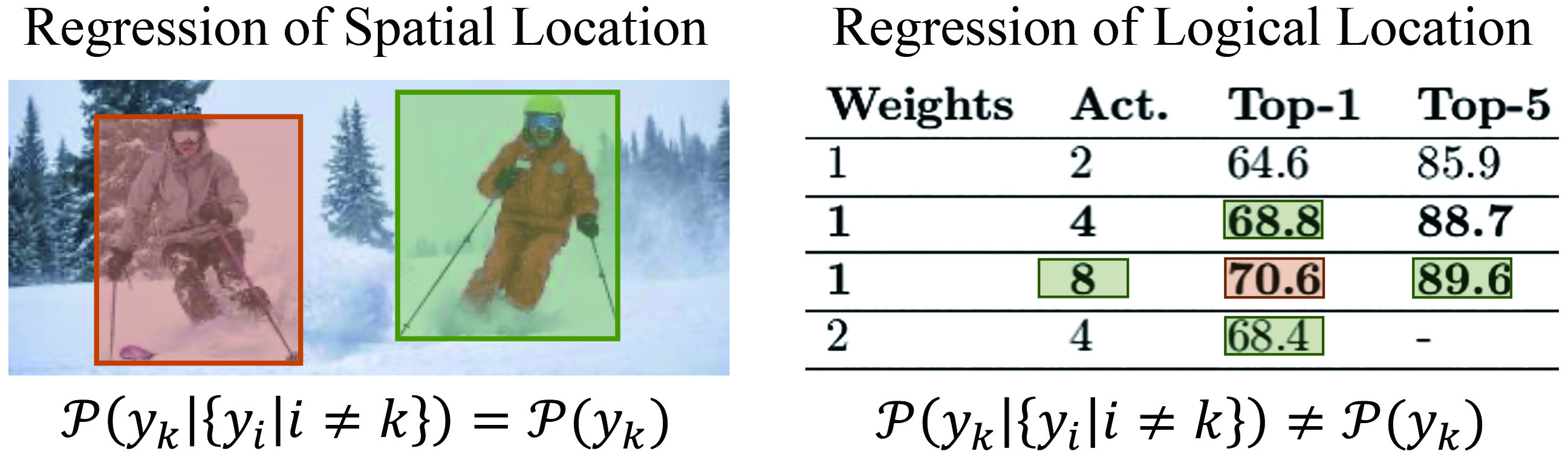}    

\caption{A comparison between the usual regression (left) and the logical location regression (right). The typical regression hypothesis is that different targets are independently distributed. However, dependencies exist between logical indices, e.g., the logical location of the cell `70.6' is constrained by those of the four surrounding cells.}
\label{fig:log}
\end{figure}

In fact, logical coordinates are well-defined machine-understandable representations of table structures, which are complete to reconstruct tables, as depicted in Figure \ref{fig:type} (c). Recently, work arises which focuses on exploring the logical locations of table cells \cite{xue2021tgrnet}. However, the method predicts logical locations by ordinal classification and does not account for the natural dependencies between logical locations. For example, the design of a table itself is from top to bottom, left to right, causing the logical location of cells to be interdependent. This nature of logical locations is sketched in Figure \ref{fig:log}. Furthermore, the work lacks a comprehensive comparison among various TSR paradigms.

Aiming at breaking the limitations of existing methods, we propose \textbf{LO}gical Location \textbf{RE}gression Network (LORE for abbreviation), a conceptually simpler and more effective TSR framework. It first locates table cells on the input image, and then predicts the logical locations along with the spatial locations of cells. To better model the dependencies and constraints between logical locations, a cascade regression framework is adopted, combined with the inter-cell and intra-cell supervisions. The inference of LORE is a parallel network forward-pass, without any efforts in complicated post-processings or sequential decoding strategies.

We evaluate LORE on a wide range of benchmarks against TSR methods of different paradigms. Experiments show that LORE is highly competitive and outperforms previous state-of-the-art methods. Specifically, LORE surpasses other logical location prediction methods by a large margin. Moreover, the adjacency relations and markup sequences derived from the prediction of LORE are of higher quality, which demonstrates that LORE covers the capacity of the models trained under other TSR paradigms.

Our main contributions can be summarized as follows:
\begin{itemize}
    \item We propose to model TSR as the logical location regression problem and design LORE, a new TSR framework which captures dependencies and constraints between logical locations of cells, and predicts the logical locations along with the spatial locations.
    
    \item We empirically demonstrate that the logical location regression paradigm is highly effective and covers the abilities of previous TSR paradigms, such as predicting adjacency relations and generating markup sequences.
    
    \item LORE provides a hands-off way to apply an effective TSR model, by removing the effort for designing post-processings and decoding strategies. The code is available to support further investigations on TSR. 
\end{itemize}

\section{Related Work}
Early works \cite{schreiber2017deepdesrt, siddiqui2019deeptabstr} introduce segmentation or detection frameworks to locate and extract splitting lines of table rows and columns. Subsequently, they reconstruct the table structure by empirically grouping the cell boxes with pre-defined rules. These models would suffer from tables with spanning cells or distortions. The latest baselines \cite{long2021parsing, smock2022pubtables, zhang2022split} tackle this problem by well-designed detectors or attention-based merging modules to obtain more accurate cell boundaries and merging results.  However, they either are tailored for the certain type of datasets or require customized processings to recover table structures, and thus can hardly be generalized. So there arise models focusing on directly predicting the table structures with neural networks.

\subsection{TSR as Cell Adjacency Exploring}

\citet{chi2019complicated} proposes to model table cells as text segmentation regions and exploit the relationships between cell pairs. Precisely, it applies graph neural networks \cite{kipf2016semi} to classify pairs of detected cells into horizontal, vertical and unrelated relations. Following this work, there are models devoted to improving the relationship classification by using elaborated neural networks and adding multi-modal features \cite{qasim2019rethinking, raja2020table, raja2022visual, liu2021show, liu2021neural}. However, there is still a gap between the set of relation triplets and the global table structure. Complex graph optimization algorithms or pre-defined post-processings are needed to recover the tables.

\subsection{TSR as Markup Sequence Generation}
\citet{li-etal-2020-tablebank, zhong2020image, ye2021pingan} make the pioneering attempts to solve the TSR problem in an end-to-end way. They employ sequence decoders to generate tags of markup language that represent table structures. However, the models are supposed to learn the markup grammar with noisy labels, resulting in the methods being difficult to train and requiring a much larger number of training samples than other paradigms. Besides, these models are time-consuming owing to the sequential decoding process.

\subsection{TSR as Logical Location Prediction}
\citet{xue2021tgrnet} propose to perform ordinal classification of logical indices on each detected cell for TSR, which is close to our approach. The model utilizes graph neural networks to classify detected cells into the  corresponding logical locations, while it ignores the dependencies and constraints among logical locations of cells. Besides, the model is only evaluated on a few datasets and not against the strong TSR baselines. 

\begin{figure*}[t]
  \centering
\includegraphics[width=1.98\columnwidth]{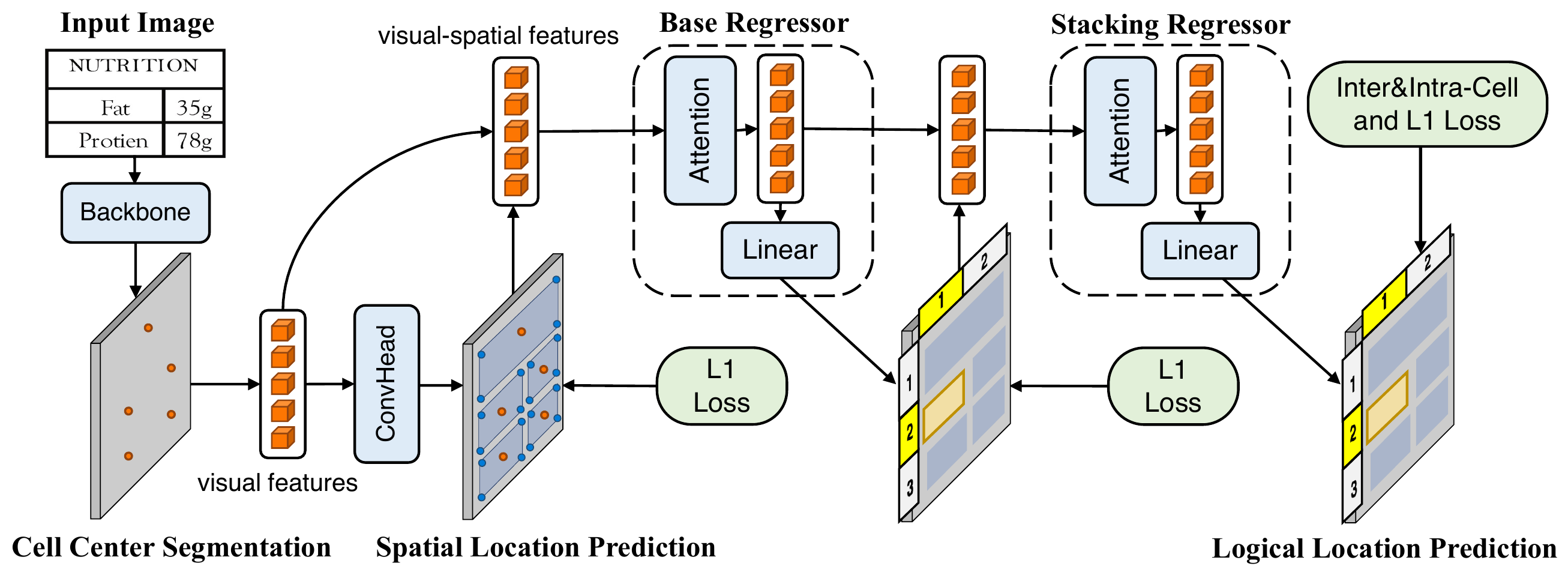}    

\caption{An illustration of LORE. It first locates table cells in the input image by key point segmentation. Then the logical locations are predicted along with the spatial locations. The cascading regressors and the inter-cell and intra-cell supervisions are employed to better model the dependencies and constraints between logical locations. }
\label{fig:model}
\end{figure*}

\section{Problem Definition}
In this paper, we consider the TSR problem as the spatial and logical location regression task. Specifically, for an input image of the table, similar to a detector, a set of table cells $\{O_1,O_2,...,O_N\}$ are predicted as their logical locations $\{l_1, l_2, ..., l_N\}$, along with the spatial locations $\{B_1, B_2, ..., B_N\}$, where $l_i = (r_s^{(i)}, r_e^{(i)}, c_s^{(i)}, c_e^{(i)})$ standing for the starting-row, ending-row, starting-column and ending-column, $B_i = \{(x_k^{(i)}, y_k^{(i)})\}_{k=1,2,3,4}$ standing for the four corner points of the $i$-th cell and $N$ is the number of cells in the image.

With the predicted table cells represented by their spatial and logical locations, the table in the image can be converted into machine-understandable formats, such as relational databases. Besides, the adjacency matrices and the markup sequences of tables can be directly derived from their logical coordinates with well-defined transformations rather than heuristic rules (See supplementary section 1).

\section{Methodology}
This section elaborates on our proposed LORE, a TSR framework regressing the spatial and logical locations of cells. As illustrated in Figure \ref{fig:model}, it employs a CNN backbone to extract visual features of table cells from the input image. Then the spatial and logical locations of cells are predicted by two regression heads. We specially leverage the cascading regressors and employ inter-cell and intra-cell supervisions to model the dependencies and constraints between logical locations. The following subsections specify these crucial components respectively.

\subsection{Table Cell Features Preparation}
In order to streamline the joint prediction of spatial and logical locations, we employ a key point segmentation network \cite{zhou2019objects,long2021parsing} as the feature extractor and model each table cell in the image as its center point.

For an input image of width $W$ and height $H$, the network produces a feature map $f \in \mathbb{R}^{\frac{W}{R} \times \frac{H}{R}\times d} $ and a cell center heatmap $\widehat{Y} \in [0,1]^{\frac{W}{R} \times \frac{H}{R}} $, where $R$, $d$ are the output stride and hidden size; $ \widehat{Y}_{x,y} = 1 $ corresponds to a detected cell center, while $ \widehat{Y}_{x,y} = 0 $ refers to the background. 

In the subsequent modules, the CNN features $\{f^{(1)}, f^{(2)},..., f^{(N)}\}$ at detected cell centers $\{\hat{p}^{(1)}, \hat{p}^{(2)}, ..., \hat{p}^{(N)}\}$ are considered as the representations of table cells.

\subsection{Spatial Location Regression} 
 
We choose to predict the four corner points rather than the rectangle bounding box to better deal with the inclines and distortions of tables in the wild. For spatial locations, the features of the backbone $f$ are passed through a $3 \times 3$ convolution, ReLU and another $1 \times 1$ convolution to get the prediction $\{\hat{B}^{(1)}, \hat{B}^{(2)}, ..., \hat{B}^{(N)}\}$ on centers $\{\hat{p}^{(1)}, \hat{p}^{(2)}, ..., \hat{p}^{(N)}\}$, where $\hat{B}^{(i)} = \{(\hat{x}_k^{(i)}, \hat{y}_k^{(i)})\}_{k=1,2,3,4}$.

\subsection{Logical Location Regression}
As dense dependencies and constraints exist between the logical locations of table cells, it is rather challenging to learn the logical coordinates from the visual features of cell centers alone. The cascading regressors with inter-cell and intra-cell supervisions are leveraged to explicitly model the logical relations between cells.

\subsubsection{Base Regressor}
To better model the logical relations from images, the visual features are first combined with the spatial information. Specifically, the features of the predicted corner points of the cells are computed as the sum of their visual features and 2-dimensional position embeddings:

\begin{equation}
    \widetilde{f}_{(\hat{x}_k^{(i)},\hat{y}_k^{(i)},:)} = 
f_{(\hat{x}_k^{(i)},\hat{y}_k^{(i)},:)} + PE(\hat{x}_k^{(i)},\hat{y}_k^{(i)}),
\end{equation}
where $PE$ refers to the 2-dimensional position embedding function \cite{xu2020layoutlm, xu2020layoutlmv2}. Then the features of the four corner points are added to the center features $f^{(i)}$ to enhance the representation of each predicted cell center $\hat{p}^{(i)}$ as:
\begin{equation}
    h^{(i)} = f^{(i)} + \sum_{k=1}^4 w_k \widetilde{f}_{(\hat{x}_k^{(i)},\hat{y}_k^{(i)},:)},
\end{equation}
where $[w_1, w_2, w_3, w_4]$ are learnable parameters.

Then the message-passing and aggregating networks are adopted to incorporate the interaction between the visual-spatial features of cells:
\begin{equation}
    \{ \widetilde{h}^{(i)}\}_{i = 1,2,...,N} = {\rm \textbf{Self\-Attention}}(\{ h^{(i)}\}_{i = 1,2,...,N}).
\end{equation} 

We use the self-attention mechanism \cite{vaswani2017attention} in LORE to avoid making additional assumptions about the distribution of table structure, rather than graph neural networks employed by previous methods \cite{qasim2019rethinking, xue2021tgrnet}, which will be further discussed in experiments.

The prediction of the base regressor is then computed by a linear layer with the ReLU activation from $\{ \widetilde{h}^{(i)}\}_{i = 1,2,...,N} $ as $\hat{l}^{(i)} = (\hat{r}_{s}^{(i)}, \hat{r}_{e}^{(i)}, \hat{c}_{s}^{(i)}, \hat{c}_{e}^{(i)})$.

\subsubsection{Stacking Regressor}
Although the base regressor encodes the relationships between visual-spatial features of cells, the logical locations of each cell are still predicted individually. To better capture the dependencies and constraints among logical locations, a stacking regressor is employed to look again at the prediction of the base regressor. Specifically, the enhanced features $\boldsymbol{\widetilde{h}}$ and the logical location prediction of the base regressor $\hat{\boldsymbol{l}}$ are fed into a stacking regressor. The stacking regressor can be expressed as :

\begin{equation}
 \widetilde{\boldsymbol{l}} = F_s(W_s\hat{\boldsymbol{l}}+ {\boldsymbol{\widetilde{h}}}).
\end{equation}

where $W_s \in \mathbb{R}^{4\times d}$ is a learnable parameter,  $ \hat{\boldsymbol{l}} = [\hat{l}^{(1)},...,\hat{l}^{(N)}]$, $ {\boldsymbol{\widetilde{h}}} = [\widetilde{h}^{(1)},...,\widetilde{h}^{(N)}]$ and $F_s$ denotes the stacking regression function, which has the same self-attention and linear structure as the base regression function but with independent parameters. The output of the stacking regressor is $ \widetilde{{\boldsymbol{l}}} = [\widetilde{l}^{(1)},...,\widetilde{l}^{(N)}]$, and $\widetilde{l}^{(i)} = (\widetilde{r}_{s}^{(i)}, \widetilde{r}_{e}^{(i)}, \widetilde{c}_{s}^{(i)}, \widetilde{c}_{e}^{(i)})$.

At the inference stage, the results are obtained by assigning the four components of $\widetilde{l}^{(i)}$ to the nearest integers. 

\subsubsection{Inter-cell and Intra-cell Supervisions}
In order to equip the logical location regressor with a better understanding of the dependencies and constraints between logical locations, we propose the inter-cell and intra-cell supervisions, which are summarized as: 1) The logical locations of different cells should be mutually exclusive (inter-cell). 2) The logical locations of one table cell should be consistent with its spans (intra-cell).

In practice, predictions of cells that are far apart rarely contradict each other, so we only sample adjacent pairs for inter-cell supervision. More formally, the scheme of inter-cell and intra-cell losses can be expressed as:

\begin{equation}
\begin{aligned}
    L_{inter} & = \sum_{(i,j) \in A_r}max(\widetilde{r}_e^{(j)} - \widetilde{r}_s^{(i)} + 1,  0) \\
    & + \sum_{(i,j) \in A_c}max(\widetilde{c}_e^{(j)} - \widetilde{c}_s^{(i)} + 1, 0),
\end{aligned}
\end{equation}
where $A_r$ ($A_c$) are sets of ordered horizontally (vertically) adjacent pairs, i.e., for a pair of cells $(i,j) \in A_r$ ($A_c$), cell $i$ is adjacent to cell $j$ in the same row (column) and on the right of (under) cell $j$, and $\widetilde{r}_s^{(i)}$,  $\widetilde{r}_e^{(j)}$, $\widetilde{c}_s^{(i)}$, $\widetilde{c}_e^{(j)}$ are predicted logical indices of cell $i$ and cell $j$. 

\begin{equation}
\begin{aligned}
    L_{intra} & = \sum_{i \in M_r}|\widetilde{r}_s^{(i)} - \widetilde{r}_e^{(i)}  - r_s^{(i)} + r_e^{(i)}| \\ 
    & + \sum_{i \in M_c}|\widetilde{c}_s^{(i)} - \widetilde{c}_e^{(i)} - c_s^{(i)} + c_e^{(i)} |,
\end{aligned}
\end{equation}
where $M_r = \{i|r_e^{(i)} - r_s^{(i)} \neq 0\}$ and $M_c = \{i|c_e^{(i)} - c_s^{(i)} \neq 0\}$ are sets of multi-row and multi-column cells. 

Then the inter-cell and intra-cell losses (I2C) are as:
$$L_{I2C} = L_{inter} + L_{intra}.$$

The supervisions are conducted on the output $\widetilde{{\boldsymbol{l}}}$ and no extra forward-passing is required.

\subsection{Objectives}
The losses of cell center segmentation $L_{center}$ and spatial location regression $L_{spa}$ are computed following typical key point-based detection methods \cite{zhou2019objects,long2021parsing}. 

The loss of logical locations is computed for both the base regressor and the stacking regressor:
\begin{equation}
     L_{log} =  \frac{1}{N} \sum_{i=1}^N (||\hat{l}^{(i)} - l_i||_1 + ||\widetilde{l}^{(i)} - l_i||_1).
\end{equation}

The total loss of joint training is then computed by adding the losses of cell center segmentation, spatial and logical location regression along with the I2C supervisions:
\begin{equation}
     L_{LORE} = L_{center} + L_{spa} + L_{log} + L_{I2C}.
\end{equation}

\begin{table*}
    \centering
    \setlength\tabcolsep{0.437cm}
    \begin{tabular}{lccccccccccc}
    \toprule
    
    Datasets & \multicolumn{2}{c}{ICDAR-13}  & \multicolumn{2}{c}{ICDAR-19}   & \multicolumn{2}{c}{WTW} & \multicolumn{2}{c}{TG24K}   \\

     metric & F-1 & Acc & F-1 & Acc & F-1 & Acc & F-1 & Acc \\
    \midrule
    ReS2TIM & - & 17.4 & - & 13.8 & - & - & - &  - \\
    TGRNet    & 66.7 & 27.5 & 82.8 & 26.7 & 64.7 & 24.3 & 92.5 & 84.5 \\
    \midrule
    Ours        &  \underline{97.2} & \underline{86.8} &  \underline{90.6} & \underline{73.2}  & \underline{ 96.4} & \underline{82.9} & \underline{96.1} & \underline{87.9} \\
    \bottomrule
    \end{tabular}
    \caption{Comparison with the TSR methods predicting logical locations. F-1 score here is the metric for cell detection. Underlines denote the best.}
     \label{tab:logical}

\end{table*}

\begin{table*}[t]
\centering
\begin{tabular}{lccccccccccccccc|}
\toprule
Datasets & \multicolumn{3}{c}{ICDAR-13}   & \multicolumn{3}{c}{SciTSR-comp}  & \multicolumn{3}{c}{ICDAR-19}   & \multicolumn{3}{c}{WTW}     \\
 metric & P & R & F-1 & P & R & F-1 & P & R & F-1 & P & R & F-1  \\
 
\midrule
\noalign{\smallskip}

TabStrNet & 93.0 & 90.8 & 91.9  & 90.9 & 88.2 & 89.5 & 82.2  & 78.7 & 80.4 & - & - & -  \\
LGPMA    & 96.7 & 99.1 & 97.9  & 97.3 & 98.7 & 98.0 & - & - & - & - & - & - \\
TOD   & 98.0 & 97.0 & 98.0  & 97.0 & 99.0 & 98.0 & 77.0 & 76.0 & 77.0 & - & - & -  \\
FLAGNet & 97.9 & \underline{99.3} & 98.6  & 98.4 & 98.6 & 98.5 & 85.2 & 83.8 & 84.5 & 91.6  & 89.5  & 90.5   \\
NCGM & 98.4 & \underline{99.3} & 98.8  & 98.7 & 98.9 & 98.8 & 84.6 & 86.1 & 85.3 & 93.7 & 94.6 & 94.1  \\
\midrule
Ours &   \underline{99.2}  & 98.6 & \underline{98.9} & \underline{99.4} & \underline{99.2} & \underline{99.3} & \underline{87.9} & \underline{88.7} & \underline{88.3} & \underline{94.5} & \underline{95.9} & \underline{95.1} \\
\bottomrule
\end{tabular}
\caption{Comparison with the TSR methods predicting adjacency of cells. The precision, recall and F-1 score are evaluated on adjacency relationship-based metrics. Underlines denote the best.}
\label{tab:adjacency}

\end{table*}

\begin{table}
    \centering
    \setlength\tabcolsep{0.33cm}
    \begin{tabular}{lccc}
    \toprule
    Datasets & \multicolumn{1}{c}{PubTabNet}    & \multicolumn{2}{c}{TableBank}    \\

     metric  & TEDS & TEDS & BLEU \\
    \midrule
    Image2Text & - & -  & 73.8  \\
    EDD & 89.9 & 86.0  & -  \\
    
    \midrule

    Ours & \underline{98.1} & \underline{92.3} & \underline{91.1}  \\
    \bottomrule
    \end{tabular}
    \caption{Comparison with the TSR methods generating markup sequences.  Underlines denote the best.}
     \label{tab:markup}

\end{table}

\section{Experiments}

In this section, we conduct comprehensive experiments to research and answer two key questions: 1) Is the proposed LORE able to effectively predict the logical locations of table cells from input images? 2) Does the LORE framework, modeling TSR as logical location regression, overcome the limitations and cover the abilities of other paradigms?

For the first question, we compare LORE with baselines directly predicting logical locations \cite{xue2019res2tim, xue2021tgrnet}. To the best of our knowledge, these are the only two methods that focus on directly predicting the logical locations. Furthermore, we provide a detailed ablation study to validate the effectiveness of the main components. For the second question, we compare LORE with methods that model table structure as cell adjacency or markup sequence with both insights and quantitative results.

\subsection{Datasets} 

We evaluate LORE on a wide range of benchmarks, including tables in digital-born documents, i.e., ICDAR-2013 \cite{gobel2013icdar}, SciTSR-comp \cite{chi2019complicated}, PubTabNet \cite{zhong2020image}, TableBank \cite{li-etal-2020-tablebank} and TableGraph-24K \cite{xue2021tgrnet}, as well as tables from scanned documents and photos, i.e.,  ICDAR-2019 \cite{gao2019icdar} and WTW \cite{long2021parsing}. Details of datasets are available in section 2 of the supplementary. It should be noted that ICDAR-2013 provides no training data, so we extend it to the partial version for cross validation following previous works \cite{raja2020table, liu2021neural, liu2021show}. And when training LORE on the PubTabNet, we randomly choose 20,000 images from its training set for efficiency.

\subsection{Evaluation Metric}
The TSR models of different paradigms are evaluated using different metrics,  including 1) accuracy of logical locations \cite{xue2019res2tim}, 2) F-1 score of adjacency relationships between cells \cite{gobel2012methodology, gobel2013icdar}, and 3) BLEU and TEDS \cite{papineni2002bleu, zhong2020image}. We provide a detailed introduction of these metrics in section 3 of the supplementary. The accuracy of logical locations, BLEU and TEDS directly reflect the correctness of the predicted structure, while the adjacency evaluation only measures the quality of intermediate results of the structure. In our experiments, LORE is evaluated under all three types of metrics, since the logical coordinates are complete for representing table structures and can be converted into adjacency matrices and markup sequences by simple and clarified transformations (see section 1 of the supplementary material). When evaluating on TEDS, we use the non-styling text extracted from PDF files following \citet{zheng2021global}. We also report the performance of cell spatial location prediction, using the F-1 score under the IoU threshold of 0.5, following recent works \cite{raja2020table, xue2021tgrnet}.

\begin{table*}
    \centering
    \setlength\tabcolsep{0.27cm}
    \begin{tabular}{lcccccccccc}
    \toprule
    
 \multirow{2}{*}{N}   & \multicolumn{3}{c}{Objectives}  & \multirow{2}{*}{Cascade} & \multicolumn{3}{c}{Architecture}  &  \multicolumn{3}{c}{Metrics}    \\

     & $L_1$ & Inter & Intra &  &Encoder & Base & Stacking  & A-c &  A-r & Acc  \\
    
  \midrule
  1a & \checkmark & - & - & \checkmark & Attention &3 & 3 & 87.2 &  84.8 & 79.4  \\
  1b & \checkmark & \checkmark & - & \checkmark & Attention &3 & 3 & 87.6 &  86.6 & 80.2  \\
  1c &\checkmark & - & \checkmark & \checkmark & Attention &3& 3  & 89.5 & 87.1 & 81.2  \\
  1d & \checkmark & \checkmark & \checkmark & \checkmark & Attention &3 & 3  & 91.3 & 87.9 & 82.9 \\
  \midrule
   
  2a & \checkmark & \checkmark & \checkmark & \checkmark &  GNN & 3& 3 & 88.2 & 82.6 & 77.0 \\
  2b & \checkmark & \checkmark & \checkmark & - & Attention & 6 & 0  & 88.7  & 85.3 & 79.8 \\

    \bottomrule
    \end{tabular}
    \caption{Ablation study of LORE. A-c, A-r and Acc refer to the accuracy of column indices, row indices and all logical indices. All these models are trained from scratch according to the `Implementation' section.}
     \label{tab:ablation}
\end{table*}

\subsection{Implementation}
LORE is trained and evaluated on table images with the max side scaled to a fixed size of $1024$ ($512$ for SciTSR and PubTabNet) and the short side resized equally. The model is trained for 100 epochs, and the initial learning rate is chosen as $1 \times 10^{-4}$, decaying to $1 \times 10^{-5}$ and $1 \times 10^{-6}$ at the 70th and 90th epochs for all benchmarks. All the experiments are performed on the platform with 4 NVIDIA Tesla V100 GPUs. We use the DLA-34 \cite{yu2018deep} backbone, the output stride $R = 4$ and the number of channels $d = 256$. When implementing on the WTW dataset, a corner point estimation is equipped following \citet{long2021parsing}. The number of attention layers is set to 3 for both the base and the stacking regressors. We run the model 5 times and take the average performance.

\subsection{Results on Benchmarks}
First, we compare LORE with models which directly predict logical locations including Res2TIM \cite{xue2019res2tim} and TGRNet \cite{xue2021tgrnet}. We tune the model provided by \citet{xue2021tgrnet} on WTW dataset to make a thorough comparison. As shown in Table \ref{tab:logical}, LORE outperforms the previous methods remarkably. The baseline methods can only produce passable results on relatively simple benchmarks of digital-born table images from scientific articles, i.e., TableGraph-24K.

Then we compare LORE with models mining the adjacency of cells by relation-based metrics: TabStrNet \cite{raja2020table}, LGPMA \cite{qiao2021lgpma}, TOD \cite{raja2022visual}, FLAGNet \cite{liu2021show} and NCGM \cite{liu2021neural}. The adjacency relation results of LORE are derived from the output logical locations as mentioned before. The results are shown in Table \ref{tab:adjacency}. It is worth noting that LORE performs much better on challenging benchmarks such as ICDAR-2019 and WTW with scanned documents and photos. Tables in these datasets are with more spanning cells and distortions \cite{liu2021neural, long2021parsing}. Experiments demonstrate that LORE is capable of predicting adjacency relations, as by-products of regressing the logical locations.    

Finally, we evaluate LORE on the markup sequence generation scene against Image2Text \cite{li-etal-2020-tablebank} and EDD \cite{zhong2020image}, with the results also derived from the output logical locations of LORE. Specially, since the TableBank dataset does not provide the spatial locations of cells, we implement LORE trained on SciTSR (1/10 the size of TableBank) for the evaluation on it. The results are shown in Table \ref{tab:markup}. Experiment results indicate that LORE is also more effective even if LORE is trained on much fewer samples.

\subsection{Ablation Study}
To investigate how the key components of our proposed LORE contribute to the logical location regression, we conduct an intensive ablation study on the WTW dataset. Results are presented in Table \ref{tab:ablation}. First, we evaluate the effectiveness of the inter-cell loss $L_{inter}$ and the intra-cell loss $L_{intra}$, by training several models turning them on and off. According to the results in experiments 1a and 1b, we see that the inter-cell supervision improves the performance by +0.8\%Acc. And from 1a and 1c, the intra-cell supervision benefits more by +1.8\%Acc, for the reason that it makes up the message-passing and aggregating mechanism, which pays less attention to intra-cell relations than inter-cell relations according to its inter-cell nature. The combination of the two supervisions makes the best performance.

Then we evaluate the influence of model architecture, i.e., the pattern of message  aggregation and the importance of the cascade framework. In experiment 2a, we replace the self-attention encoder with a graph-attention encoder similar to graph-based TSR models \cite{qasim2019rethinking, xue2021tgrnet} with an equal amount of parameters with LORE. It causes a drop in performance consistently. The graph-based encoder only aggregates information from the top-K nearest features of each node based on Euclidean distance, which is biased for table structure. In experiment 2b, we use a single regressor of 6 layers instead of two cascading regressors of 3 layers. We can observe a performance degradation of 3.1\%Acc from 1d to 2b, showing that the cascade framework can better model the dependencies and constraints between logical locations of different cells. 

\begin{figure}[t]
    \centering
    \subfigure[Original structure] {\includegraphics[width=0.99\linewidth]{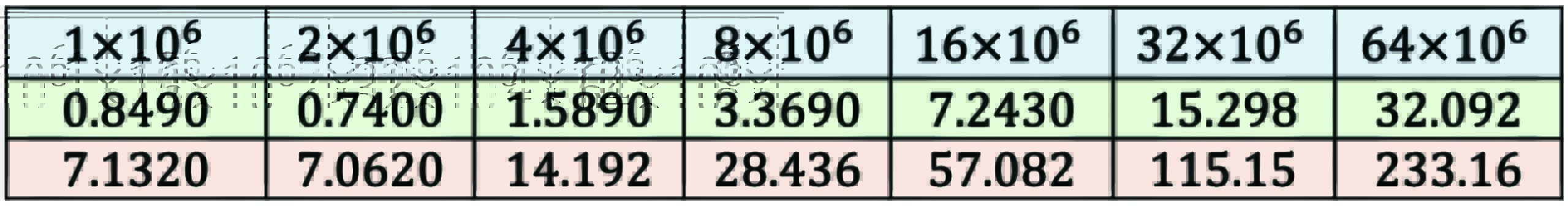}}
    
    \subfigure[Shifted structure] {\includegraphics[width=0.99\linewidth]{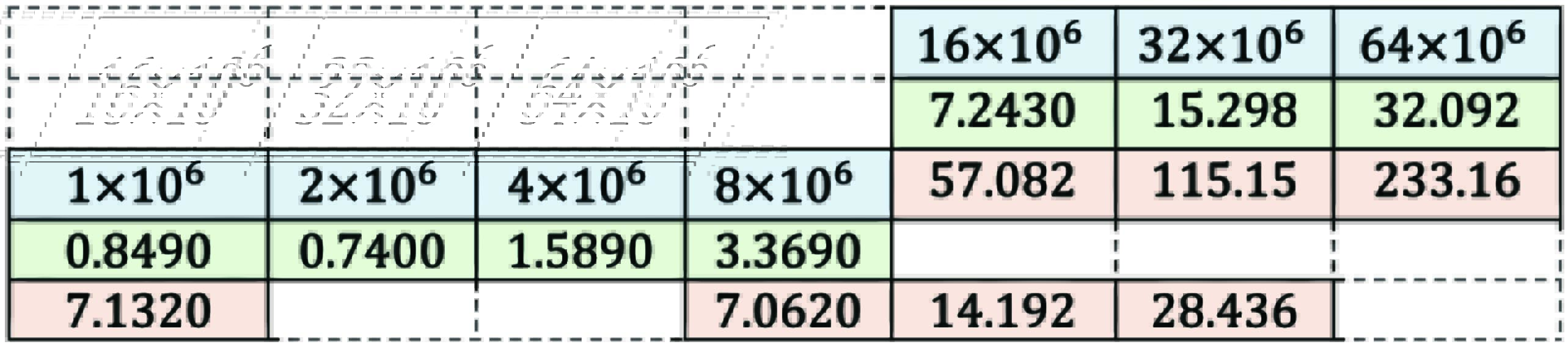}}
    
    \caption{An example of severely shifted structure. Its adjacency-relationship F-1 is 84\%, while the logical location accuracy is 43\%.}
    
    \label{fig:shift}
    
\end{figure}

\subsection{Further Comparison among Paradigms}
In this section, we further compare models of different TSR paradigms introduced before. Previous methods that predict logical locations lack a comprehensive comparison and analysis between these paradigms. We demonstrate how LORE overcomes the limitations of the adjacency-based and the markup-based methods by controlled experiments.

The adjacency of cells alone is not sufficient to represent table structures. Previous methods employ heuristic rules based on spatial locations \cite{liu2021neural} or graph optimizations \cite{qasim2019rethinking} to reconstruct the tables. However, it takes tedious modification to make the pre-defined parts compatible with datasets of different types of tables and annotations. Furthermore, the adjacency-based metrics sometimes fail to reflect the correctness of table structures, as depicted in Figure \ref{fig:shift}. Experiments are conducted to verify this argument quantitatively. We turn the linear layer of the stacking regressor of LORE into an adjacency classification layer of paired cell features and employ post-processings as in NCGM \cite{liu2021neural} to reconstruct the table. The results are in Table \ref{tab:rflogi}. Although this modified model (Adj. paradigm) achieves competitive results with state-of-the-art baselines evaluated on adjacency-based metrics, the accuracy of logical locations obtained from heuristic rules decreases obviously compared to LORE (Log. paradigm), especially on WTW, which contains more spanning cells and distortions.

\begin{figure}[t]
  \centering
    \subfigure[Attention activation of the base regressor] {\includegraphics[width=0.97\linewidth]{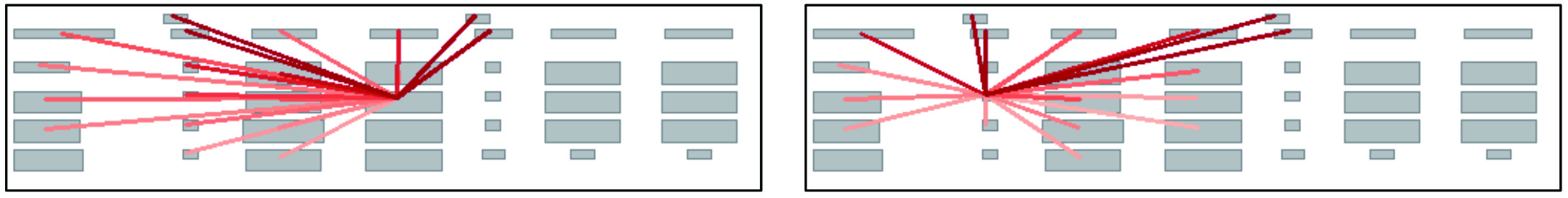}}
    
    \subfigure[Attention activation of the stacking regressor] {\includegraphics[width=0.97\linewidth]{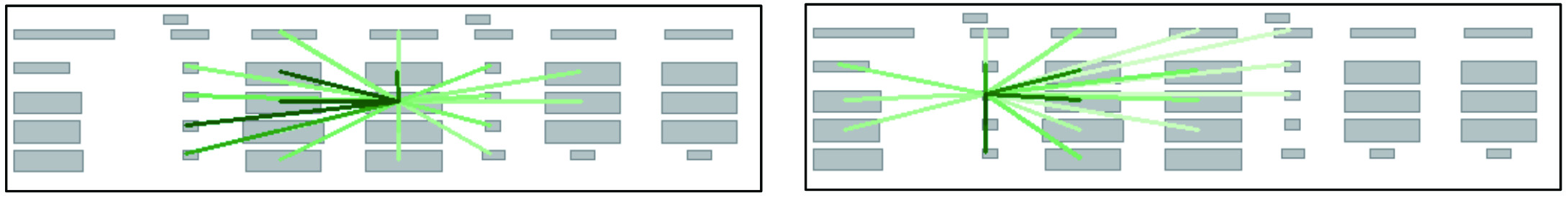}}
    
     \subfigure[Attention activation of the non-cascade regressor] {\includegraphics[width=0.97\linewidth]{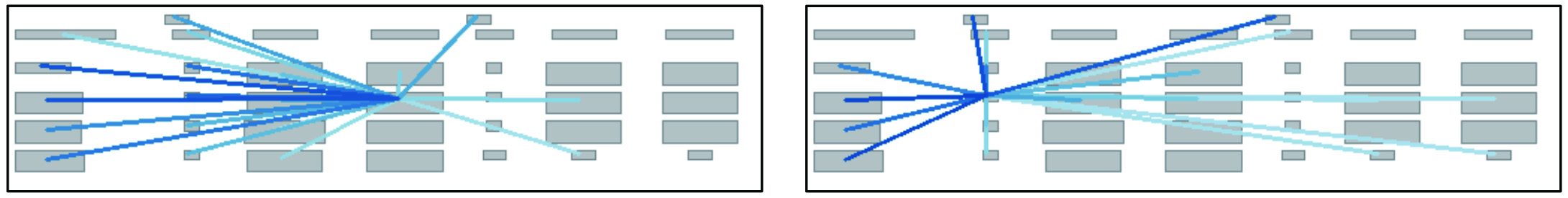}}
    
    \caption{Visualization of the self-attention weights in the cascade and non-cascade regressors for two table cells. Text masks represent table cells and only top-20 weights are visualized for clarity.}

    
    \label{fig:att}
\end{figure}

\begin{table}
    \centering
    \setlength \tabcolsep{0.19cm}
    \begin{tabular}{lccccccc}
    \toprule
    \multirow{2}{*}{Data} & \multirow{2}{*}{Paradigm} & \multicolumn{3}{c}{Adj. Metrics}    & \multicolumn{3}{c}{Log. Metrics} \\

    & & P & R & F-1 &   A-all   & A-sp \\
    \midrule
    \multirow{2}{*}{Sci-c} & Adj. & 98.6 & 98.9 & 98.7 & 94.7 & 63.5\\
    & Log. & 99.4 & 99.2  &  99.3  & 97.3 & 87.7 \\
    \midrule
    \multirow{2}{*}{WTW} &Adj. & 95.0 & 93.7  &  94.3  & 51.9 & 20.2 \\
    & Log. & 94.5 & 95.9  &  95.1  & 82.9 & 63.8\\
  
    \bottomrule
    \end{tabular}
    \caption{Evaluation results of the adjacency and the logical location paradigms. A-all and A-sp refer to the logical location accuracy of all cells and spanning cells (more than one row/column). Sci-c denotes SciTSR-comp. }
     \label{tab:rflogi}

\end{table}

The markup-sequence-based models leverage image encoders and sequence decoders to predict the label sequences. Since the markup language has plenty of control sequences formatting styles, they can be viewed as noise in labels and impede model training \cite{xue2021tgrnet}. It requires much more training samples and computational costs. As shown in Table \ref{tab:consum}, the number of training samples of the EDD model on the PubTabNet dataset is more than ten times larger than that of LORE. Besides, the inference process is rather time-consuming (See Table \ref{tab:consum}) due to the sequential decoding pattern, while models of other paradigms compute for each cell in parallel. The average inference time is computed from the validation set of PubTabNet with the images resized to $1280\times1280$ for both models.

\subsection{Further Analysis on Cascade Regressors}
We conduct experiments to investigate the effect of the cascade framework on the prediction of logical coordinates. In Figure \ref{fig:att}, we visualize the attention maps of the last encoder layer of the cascade/single regressors of two cells, i.e., the models 1d and 2b in Table \ref{tab:ablation}. In the cascade framework, the base regressor in Figure \ref{fig:att} (a) focuses on the heading cells (upper or left) to compute logical locations. While the stacking regressor in Figure \ref{fig:att} (b) pays more attention to the surrounding cells to discover finer dependencies among logical locations and make sure the prediction is subject to natural constraints, which is in line with human intuition when designing a table. However, the non-cascade regressor in Figure \ref{fig:att} (c) can only play a role similar to the base regressor, which leaves out important information for the prediction of logical locations.

\begin{table}
    \centering
    \setlength\tabcolsep{0.19cm}
    \begin{tabular}{lcc}
    \toprule
     & \#Train Samples    & Inference Time  \\
    \midrule
    EDD & 339000 & 14.8s\\
    LORE & 20000 & 0.45s \\
  
    \bottomrule
    \end{tabular}
    \caption{Comparison of LORE and the markup generation model EDD in terms of training samples and average inference time.}
     \label{tab:consum}

\end{table}

\begin{table}
    \centering
    \setlength\tabcolsep{0.42cm}
    \begin{tabular}{lccc}
    \toprule
     & DLA-34    & LORE  \\
    \midrule
    \#Params & 15.9  & 24.2 \\
    FLOPs & 74.6  & 75.2 \\
  
    \bottomrule
    \end{tabular}
    \caption{Computational Analysis. The units are million for the number of parameters and giga for the FLOPs. }
     \label{tab:flops}

\end{table}

\subsection{Computational Analysis}
We summarize the model size and the inference operations of LORE in Table \ref{tab:flops}, with the input images at $1024\times1024$ and the number of cells as 32.  It is observed that the complexity of LORE is at an equal level to a key point-based detector \cite{zhou2019objects} with the same backbone, showing the efficiency of LORE.

\section{Conclusions}
In summary, we propose LORE, a TSR framework that effectively regresses  the spatial locations and the logical locations of table cells from the input images. Furthermore, it models the dependencies and constraints between logical locations by employing the cascading regressors along with the inter-cell and intra-cell supervisions. LORE is straightforward to implement and achieves competitive results, without tedious post-processings or sequential decoding strategies. Experiments show that LORE outperforms state-of-the-art TSR methods under various metrics and overcomes the limitations of previous TSR paradigms.

\section{Acknowledgements}
This work is supported by the National Key R\&D Program of China (No. 2018YFC2002603), the National Natural Science Foundation of China (Grant No. 61972349), the Fundamental Research Funds for the Central Universities (No. 226-2022-00064), and Alibaba-Zhejiang University Joint Institute of Frontier Technologies.

\bibliography{submission}

\end{document}